\documentclass{article}

\usepackage{microtype}
\usepackage{graphicx}
\usepackage{subfigure}
\usepackage{booktabs} % for professional tables

\usepackage{tabularx}
\usepackage{multirow}
\usepackage{array}
\usepackage{makecell}
\usepackage{pdflscape}
\newcolumntype{P}[1]{>{\centering\arraybackslash}p{#1}}
\newcolumntype{M}[1]{>{\centering\arraybackslash}m{#1}}

\usepackage{hyperref}

\usepackage[accepted]{icml2026}
\makeatletter
\renewcommand{\Notice@String}{}
\makeatother

\usepackage{amsmath}
\usepackage{amssymb}
\usepackage{mathtools}
\usepackage{amsthm}

\usepackage[capitalize,noabbrev]{cleveref}

\theoremstyle{plain}

\theoremstyle{definition}

\theoremstyle{remark}

\usepackage[textsize=tiny]{todonotes}

\icmltitlerunning{Within-Model vs Between-Prompt Variability in Large Language Models for Creative Tasks}

\begin{document}

\twocolumn[
\icmltitle{Within-Model vs Between-Prompt Variability in Large Language Models for Creative Tasks}

% unless you've provided the [accepted] option to the icml2026



\begin{icmlauthorlist}
\icmlauthor{Jennifer Haase}{huberlin,WI}
\icmlauthor{Jana Gonnermann-Müller}{zib,WI}
\icmlauthor{Paul H. P. Hanel}{ess}
\icmlauthor{Nicolas Leins}{zib}
\icmlauthor{Thomas Kosch}{huberlin,WI}
\icmlauthor{Jan Mendling}{huberlin,WI}
\icmlauthor{Sebastian Pokutta}{zib,TU}
\end{icmlauthorlist}

\icmlaffiliation{huberlin}{HU Berlin, Berlin, Germany}
\icmlaffiliation{zib}{Zuse Institute Berlin, Berlin, Germany}
\icmlaffiliation{WI}{Weizenbaum Institute for the Networked Society, Berlin, Germany}
\icmlaffiliation{ess}{University of Essex, Colchester, UK}
\icmlaffiliation{TU}{TU Berlin, Berlin, Germany}

\icmlcorrespondingauthor{Jennifer Haase}{jennifer.haase@hu-berlin.de}

\icmlkeywords{Large Language Models, Prompt Engineering, Stochasticity, Creativity, Variability, Human-AI Interaction}

\vskip 0.3in
\let\thefootnote\relax\footnotemark
]

\printAffiliationsAndNotice{}

\begin{abstract}
    How much of LLM output variance is explained by prompts versus model choice versus stochasticity through sampling? We answer this by evaluating $12$ LLMs on $10$ creativity prompts with $100$ samples each ($N=12{,}000$). For output quality (originality), prompts explain $36.43\%$ of variance, comparable to model choice ($40.94\%$). But for output quantity (fluency), model choice ($51.25\%$) and within-LLM variance ($33.70\%$) dominate, with prompts explaining only $4.22\%$. Prompts are powerful levers for steering output quality, but given the substantial within-LLM variance ($10$--$34\%$),
    single-sample evaluations risk conflating sampling noise with genuine prompt or model effects.
\end{abstract}

\section{Introduction}
\label{sec:introduction}

Prompt engineering assumes that natural-language instructions steer large language models (LLMs) toward predictable, optimized outputs. This \emph{prompt-as-design} paradigm treats prompt formulation as the primary driver of model behavior, implicitly assuming that LLMs function as near-deterministic mappings from inputs to outputs. Consequently, most LLM evaluation work selects a prompt $x$, run the model once, and interprets the resulting output $y$ as the model's behavior under that prompt \citep{shiLargeLanguageModels2023, zhuPromptRobustEvaluatingRobustness2024, caoWorstPromptPerformance2024}. Variability induced by sampling from the underlying conditional distribution $p_\theta(y \mid x)$ is rarely measured and typically ignored.

This paradigm stands in contrast to the probabilistic nature of LLMs: formally, they implement a conditional distribution $p_\theta(y \mid x)$ over outputs. Decoding procedures such as temperature scaling or nucleus sampling yield realizations of a random variable rather than a fixed response. Even for temperature set to zero, practical inference stacks exhibit nondeterminism due to GPU parallelism and batching effects \citep{paleyesPromptVariabilityEffects2025}. This reproduction issue of exact content is of such importance that even start-ups (\url{https://thinkingmachines.ai/}) formed tackling it are flourishing. When this source of variance is ignored, it is effectively absorbed into other variance components such as model choice and prompt effects, potentially inflating the apparent magnitude of prompt-induced differences. Without quantifying sampling variance, it remains unknown whether observed effects reflect genuine prompt sensitivity, choice of model, or unmeasured stochasticity through variability.

We thus ask: \textit{How is variance in LLM outputs partitioned across different factors: prompt, model choice, and within-LLM variance?} We test this question in a setting where output variability is not noise, but a defining feature: creative idea generation. We select creative tasks as our experimental testbed for two pragmatic reasons. First, they provide a high-entropy output space where $p(y|x)$ is multi-modal; unlike mathematical or factual tasks with peaked distributions, open-ended generation requires the model to traverse a large semantic solution space, making it a sensitive stress-test for prompting-induced vs. stochasticity through sampling variability. Second, we employ the Alternate Uses Task (AUT; \citet{organisciakSemanticDistanceAutomated2023}), an established measurement for divergent (as in free-associative) thinking, which possesses a validated and fully automated evaluation pipeline \cite{organisciakOpenCreativityScoring2025}. This allows for the large-scale repeated sampling necessary to rigorously partition variance components with a scale that is prohibitive for human-based evaluation.
Our findings suggest that while prompts can steer output quality, the choice of model and within-LLM variability are too substantial to ignore. 

Our key contributions are:
\begin{itemize}
    \item Variance Decomposition of Generative LLM-Output: We provide a comprehensive large-scale partition of LLM output variance, with results for originality: prompts account for $36.43\%$, model-choice for $40.94\%$, and within-LLM variance for $10.56\%$; for fluency: prompts account for $4.22\%$, model-choice for $51.25\%$, and within-LLM variance for $33.70\%$.
    \item Identification of Prompt-Specific Mechanisms: We identify ``discriminative prompts'' (e.g., Persona) that elicit high-variance, high-quality responses from capable models, versus ``constraining prompts'' (e.g., Format constraints) that suppress diversity.
    \item Metric Robustness: We quantify the ``verbosity bias'' in automated creativity metrics (correlation of length of an idea with the originality score of the idea, with Pearson's $r=0.35$) and demonstrate that model rankings are robust to length-adjustment, confirming that high-performing models (e.g., Gemini 3 Pro) achieve superior quality through semantic content rather than mere elaboration.
\end{itemize}

\section{Related Work}
\label{sec:related}

\subsection{Prompt Engineering and Benchmarkings}
Research in prompt engineering treats instructions as optimizable parameters for controlling model behavior. Techniques range from manual heuristic refinement \citep{liuDesignGuidelinesPrompt2022} to automated optimization frameworks like INSTINCT \citep{linUseYourINSTINCT2024}, which treat the LLM as a black-box function $f(x) \to y$ and estimate instruction quality from single generations per example. The existence of such prompt optimization approaches is well-established, with frameworks like GEPA (Genetic-Pareto) demonstrating that LLMs can be employed to reflect on system behavior and drive targeted improvements through iterative mutation and Pareto-aware candidate selection, evolving robust, high-performing prompt variants with minimal evaluations \citep{agrawalGEPAReflectivePrompt2025}. Robustness benchmarks systematically perturb prompts via paraphrasing, rephrasing, or adversarial noise injection \citep{zhuPromptRobustEvaluatingRobustness2024, shiLargeLanguageModels2023, caoWorstPromptPerformance2024}, reporting dramatic performance swings despite the underlying task remaining unchanged. However, these evaluations typically fix the decoding process (e.g., $\text{temperature} = 0$, greedy decoding) and draw a \emph{single} sample per prompt variant, effectively collapsing the model's output distribution to one realization per condition. More sophisticated prompting frameworks inherit this determinism: DIPPER \citep{huDipperDiversityPrompts2025} constructs inference-time ensembles by assigning diverse prompts to identical model instances and explicitly argues that prompt variation, but not stochasticity through sampling, is the meaningful source of behavioral diversity. Together, these methodological patterns reinforce a prompt-centric view in which prompts are treated as the primary and nearly deterministic locus of control, while stochasticity through sampling variability in $p_\theta(y \mid x)$ is assumed negligible and left unmeasured. 

\subsection{Stochasticity and Inference Variability}
LLMs are probabilistic by definition, yet stochasticity is frequently treated as a nuisance to be minimized. Techniques like Self-Consistency \citep{chenUniversalSelfConsistencyLarge2023} or test-time compute scaling \citep{snellScalingLLMTestTime2024} aggregate multiple samples to improve reliability, effectively merging the distribution $p(y|x)$ into a single robust prediction. However, these approaches treat sampling diversity as a resource to be \emph{collapsed} by an aggregation operator, not as an object of measurement; they do not quantify the magnitude of sampling-induced variance nor compare it to variance induced by prompt changes. Recent work on ``LLM drift'' \citep{paleyesPromptVariabilityEffects2025} explicitly documents that setting temperature to $0$ does not guarantee fully deterministic results due to GPU parallelism, batching behavior, and non-associative floating-point operations, yet rarely quantifies this variance relative to prompt effects. 

Studies on LLM creativity suggest that model configuration substantially affects output diversity: highly aligned models may trade output-variety for consistency, while base models produce more varied outputs \citep{mohammadiCreativityHasLeft2024, westBaseModelsBeat2025}. Evidence of substantial intra-model variability has been documented in creativity evaluations: \citet{haaseHasCreativityLargelanguage2025a} conducted $100$ repeated trials per model under identical prompts and observed outputs ranging from below-average to highly original, with performance distributions spanning the full range from the bottom quartile to the top quartile. Yet when evaluation is based on $N=1$ samples per prompt, prompt-level effects and sampling variability become statistically indistinguishable: any difference observed between prompts could equally arise from stochasticity through sampling drawn from $p_\theta(y \mid x)$. The relative contribution of within-LLM versus between-prompt variance remains unknown.

\subsection{Methodological Gap: The Sampling Fallacy}
The experimental design of most prompting studies mirrors a known bias in experimental psychology termed the ``fixed-effect fallacy'' \citep{juddTreatingStimuliRandom2012}, where stochasticity components (e.g., stimuli or samples) are incorrectly treated as fixed factors. In the LLM context, taking a single generation as representative of a prompt's distribution commits a similar error. If $p_\theta(y \mid x)$ exhibits substantial spread, any single realization may deviate considerably from the central tendency.

The core problem \citep{juddExperimentsMoreOne2017, juddTreatingStimuliRandom2012} is a mismatch between the data-generating process and the analytic model. When multiple random factors contribute, collapsing one factor attributes its variance to others. Standard ANOVA procedures assume all variance sources are modeled; when ignored, variance is absorbed by other factors, inflating effect sizes. Single-sample evaluation conflates prompt effects with sampling variability, potentially overestimating prompt effects.
 
Our repeated-sampling design treats stochasticity as a random factor and models multiple sources of variation using linear mixed-effects models. By decomposing variance into prompt, between-model, and within-LLM components, we test whether prompt wording constitutes a dominant causal driver of output differences.

\section{Methodology}
\label{sec:method}

\begin{table*}[ht]
    \centering
    \caption{Prompting Strategies (P1--P10) Used in This Study}
    \label{tab:prompting-strategies}
    \begin{tabular}{lp{0.17\linewidth}p{0.47\linewidth}p{0.18\linewidth}}
    \toprule
    \textbf{ID} & \textbf{Strategy} & \textbf{Description} & \textbf{References} \\
    \midrule
    
    \multicolumn{4}{l}{\textbf{Baseline}} \\
    \midrule
    
    P1 & Direct Instruction & 
    Clear task with explicit output format; based on default AUT instruction. & \citet{organisciakSemanticDistanceAutomated2023}\\
    
    \midrule
    \multicolumn{4}{l}{\textbf{Strategic prompts}} \\
    \midrule
    
    P2 & One-Shot Example & 
    Example-Based: Includes a single concrete example to anchor the model's response style. & \citet{chenUnleashingPotentialPrompt2025, linUseYourINSTINCT2024}\\
    \addlinespace
    
    P3 & Heuristic Prompt & 
    Domain Exploration: Provides mental heuristics and creativity-stimulating cues (e.g., cross-domain thinking). & \citet{chenUnleashingPotentialPrompt2025,zhouLLMsLearnTask2024}\\
    \addlinespace
    
    P4 & Anticipatory Prompt & 
    Avoidance Instructions: Preempts common errors by instructing what \emph{not} to do (e.g., avoid generic ideas). & \citet{wangDevilsAdvocateAnticipatory2024}\\
    \addlinespace
    
    P5 & Zero-Shot CoT & 
    Chain-of-Thought (CoT): Encourages reasoning before answering by embedding a multi-step analytical structure. & \citet{chenUnleashingPotentialPrompt2025}\\
    \addlinespace
    
    P6 & Creative Persona & 
    Persona description: Assigns a high-creative identity to steer style and ambition toward original output. & \citet{oleaEvaluatingPersonaPrompting2024, zhengWhenHelpfulAssistant2024}\\
    
    \midrule
    \multicolumn{4}{l}{\textbf{Minor prompt variations}} \\
    \midrule
    
    P7 & Phrasing Change & 
    Minor Phrasing: Surface edits to the baseline prompt to test sensitivity to minor semantic rewordings. & \citet{salinasButterflyEffectAltering2024}\\
    \addlinespace
    
    P8 & Formatting Tweak & 
    Structural Constraint: Imposes strict output constraints (e.g., ``no titles or colons'') to test structural impact. & \citet{heDoesPromptFormatting2024, salinasButterflyEffectAltering2024}\\
    \addlinespace
    
    P9 & Information Order & 
    Information order: Reorders elements but keep the same information as in the baseline prompt. & \citet{salinasButterflyEffectAltering2024}\\
    \addlinespace
    
    P10 & Random Errors & 
    Typo-robustness: Injects typographical and syntactic noise to assess model error-tolerance. & \citet{salinasButterflyEffectAltering2024}\\
    \bottomrule
    \end{tabular}   
    \vspace{0.5em}

    \textbf{Note:} All prompts concluded with identical JSON formatting instructions for consistency. Full prompt texts available in Appendix Table~\ref{tab:prompting-strategies-full} and \ref{tab:prompting-strategies_minor-full}.
\end{table*}

\subsection{Stimuli: Prompt Design}
\label{sec:method:stimuli}

We designed one baseline prompt following the AUT instructions \citep{christensenAlternateUsesTest1960}, which asks for alternative uses of an every-day object, for which we chose ``plastic bottle''. Thus, the basic prompt is ``Think of all kinds of things you could do with a plastic bottle''. Additionally, we added five prompting strategies based on established prompting frameworks: providing a one-shot example, heuristic prompt, anticipating unwanted output, zero-shot chain-of-thought, and persona-based.  \citep{linUseYourINSTINCT2024}. Further, we added four prompts with minor changes: paraphrasing, formatting tweak, changing order of information and adding random spelling errors (see Table~\ref{tab:prompting-strategies} for an overview, full prompt texts are provided in the Appendix Tables~\ref{tab:prompting-strategies-full} and~\ref{tab:prompting-strategies_minor-full}).

%\subsection{Models}
%\label{sec:method:models}

%We tested 12 contemporary large language models: Claude Sonnet 4.5, DeepSeek V3.2, GPT 5.1, GPT 5.2, GPT OSS 120B, Gemini 3 Pro, Gemma 3 27B, Grok 4.1, Llama 3.3 70B, Mistral Nemo, Qwen 3 235B, and Qwen 3 235B Thinking. This selection spans both proprietary and open-source models to reflect the diversity of the current LLM ecosystem (cf. Table~\ref{tab:models}).

\subsection{Inference Configuration}
\label{sec:method:inference}

In deployed settings, users do not interact with model weights in isolation, instead they interact with a model-as-service that bundles architecture, training, alignment, default decoding parameters, and system prompts into a single interface. Our design reflects this reality: proprietary LLMs were accessed via their official provider APIs, while open-weights models were hosted on local infrastructure using the Ollama runtime environment.

We employed default hyperparameter settings as defined by each provider. This choice reflects three considerations: (1) standardizing parameters creates false equivalence since temperature effects vary across models \citep{peeperkornTemperatureCreativityParameter2024}; (2) defaults represent vendor-optimized configurations; (3) using defaults maximizes ecological validity. Variance attributed to the model factor reflects the combined effect of architecture, alignment, and provider-specific defaults (see Limitations). 

\subsection{Procedure and Data Collection}
\label{sec:method:procedure}

For each model (cf. Table~\ref{tab:models}) $\times$ prompt pair, we generated $100$ independent outputs, totaling $12{,}000$ generations. All data were collected in December $2025$ via the respective provider APIs. No additional system prompts or few-shot context were provided beyond the specific stimuli described in Table~\ref{tab:prompting-strategies}. Responses were required to adhere to JSON formatting. Of the $12{,}000$ collected responses, $130$ ($1.1\%$) could not be scored due to timeouts in the automated creativity scoring service and were excluded, yielding $N=11{,}870$ valid complete cases. These were parsed into $566{,}916$ individual ideas for analysis.

\subsection{Evaluation \& Analysis}

\label{sec:method:evaluation}

We employed a multi-faceted evaluation strategy combining automated scoring and structural text analyses.

\paragraph{Data Preprocessing.}
We evaluated outputs using an automated AUT scoring system,  which assigns \textit{originality} scores on a $1$--$5$ scale \cite{organisciakOpenCreativityScoring2025}. Higher scores indicate greater originality and unconventionality of the idea. Response-level originality was computed as the mean originality score across all valid ideas in a response.

\textit{Fluency} was defined as the count of valid ideas generated per response. To ensure validity, we applied an automated parsing and filtering pipeline: (a) extraction from the required JSON structure, (b) removal of empty strings or formatting artifacts.

\paragraph{Variance Decomposition.} We partition total variance using a linear mixed-effects model. For a creativity score $Y_{mpr}$ (model $m$, prompt $p$, run $r$):
\begin{equation}
Y_{mpr} = \mu + \alpha_m + \beta_p + (\alpha\beta)_{mp} + \epsilon_{mpr}
\end{equation}
where $\mu$ is the grand mean, $\alpha_m \sim N(0, \sigma^2_{\text{model}})$ is the random intercept for LLMs, $\beta_p \sim N(0, \sigma^2_{\text{prompt}})$ is the random intercept for prompts, $(\alpha\beta)_{mp}$ is the model $\times$ prompt interaction, and $\epsilon_{mpr} \sim N(0, \sigma^2_{\text{within}})$ is the within-LLM error capturing stochasticity through sampling variance. Parameters were estimated using Restricted Maximum Likelihood (REML). Variance components are reported as Intra-class Correlation Coefficients (ICC), representing the proportion of total variance explained by each factor \citep{Bliese2000WithinGroup}. We computed $95\%$ confidence intervals via stratified bootstrap resampling ($1{,}000$ iterations) within model--prompt cells.

\paragraph{Content Analysis (Thematic \& Structural).} To characterize the qualitative differences in output, we analyzed the content on two dimensions:
\begin{itemize}
    \item \textit{Thematic Categorization.} We developed a keyword-based taxonomy tailored to the ``Plastic Bottle'' AUT task, consisting of $105$ distinct keywords mapped to $10$ functional domains (e.g., \textit{Gardening} $\leftarrow$ \{\texttt{planter}, \texttt{greenhouse}, \ldots\}; \textit{Survival} $\leftarrow$ \{\texttt{raft}, \texttt{filter}, \ldots\}). Ideas were classified using a deterministic waterfall matching algorithm: an idea was assigned to the first category where a case-insensitive keyword substring match was found. Ideas containing no keywords were labeled ``Miscellaneous''\cite{grimmerTextDataPromise2013}. To quantify the breadth of a model's semantic range, we computed the \textit{Thematic Entropy} (Shannon Entropy) of its category distribution: $H(M) = -\sum_{c \in C} p(c|M) \ln p(c|M)$, where $p(c|M)$ is the proportion of model $M$'s ideas falling into category $c$ \citep{linDivergenceMeasuresBased1991}.
    \item \textit{Repetition \& Uniqueness Analysis.} To assess the effective diversity of the generated output distributions, we performed a global repetition analysis on the full corpus of $N=584{,}205$ parsed ideas. We applied a standard string normalization pipeline (lowercasing, removal of trailing punctuation, and whitespace trimming) to map surface variations to canonical forms. We then computed two metrics: (1) \textit{Global Uniqueness}, defined as the proportion of ideas that appeared exactly once across the entire dataset, serving as a measure of the total semantic space covered; and (2) \textit{Intra-Model Repetition Rate}, the percentage of a model's outputs that were duplicates of its own previous generations. This metric serves as a proxy for \textit{mode collapse}, differentiating models that explore the tail of the distribution from those that greedily converge on a narrow set of high-probability sequences.
    \item \textit{Structural Profiling.} We systematically analyzed low-level text features to identify alignment fingerprints. For every idea, we extracted: (1) \textit{Verbosity} (character count), (2) \textit{Punctuation Style} (presence of colons vs. hyphens/dashes), and (3) \textit{Numerical Density} (presence of digits using the regex `\d`). These features were aggregated to measure the rigidity of a model's output format independent of the prompt.
\end{itemize}

\begin{table}[h]
    \centering
    \caption{Large Language Models used in this study}
    \label{tab:models}
    \begin{tabular}{lll}
    \toprule
    \textbf{Model Name} & \textbf{Size} & \textbf{Provider} \\
    \midrule
    Claude Sonnet 4.5$^{\dagger}$& Large & Anthropic \\
    DeepSeek Reasoner V3.2$^{\dagger}$& Large & DeepSeek \\
    GPT 5.1$^{\dagger}$& Large & OpenAI \\
    GPT 5.2$^{\dagger}$& Large & OpenAI \\
    GPT OSS 120B & $120$B & OpenAI \\
    Gemini 3 Pro$^{\dagger}$& Large & Google \\
    Gemma 3 27B & $27$B & Google \\
    Grok 4.1 & Large & xAI \\
    Llama 3.3 70B & $70$B & Meta \\
    Mistral Nemo & $12$B & Mistral \\
    Qwen 3 235B & $235$B & Alibaba \\
    Qwen 3 235B Thinking$^{\dagger}$ & $235$B & Alibaba \\
    \bottomrule
    \end{tabular}
    \vspace{0.5em}
    
    \textbf{Note:} $^{\dagger}$Reasoning/thinking model.
    \end{table}

\section{Results}
\label{sec:results}

We evaluate two outcome metrics: \textit{originality} (semantic novelty of ideas, scored 1--5) and \textit{fluency} (number of ideas generated). We partition variance into three sources: \textit{model} (systematic differences between LLMs), \textit{prompt} (systematic differences between prompting strategies), and \textit{within-LLM variance} (variability observed when querying the same model with the same prompt repeatedly). Following the core variance decomposition, we investigate the mechanisms underlying model and prompt effects.

\subsection{Variance Decomposition}
\label{sec:results:variance}

\begin{figure}[h]
    \centering
    \includegraphics[width=\columnwidth]{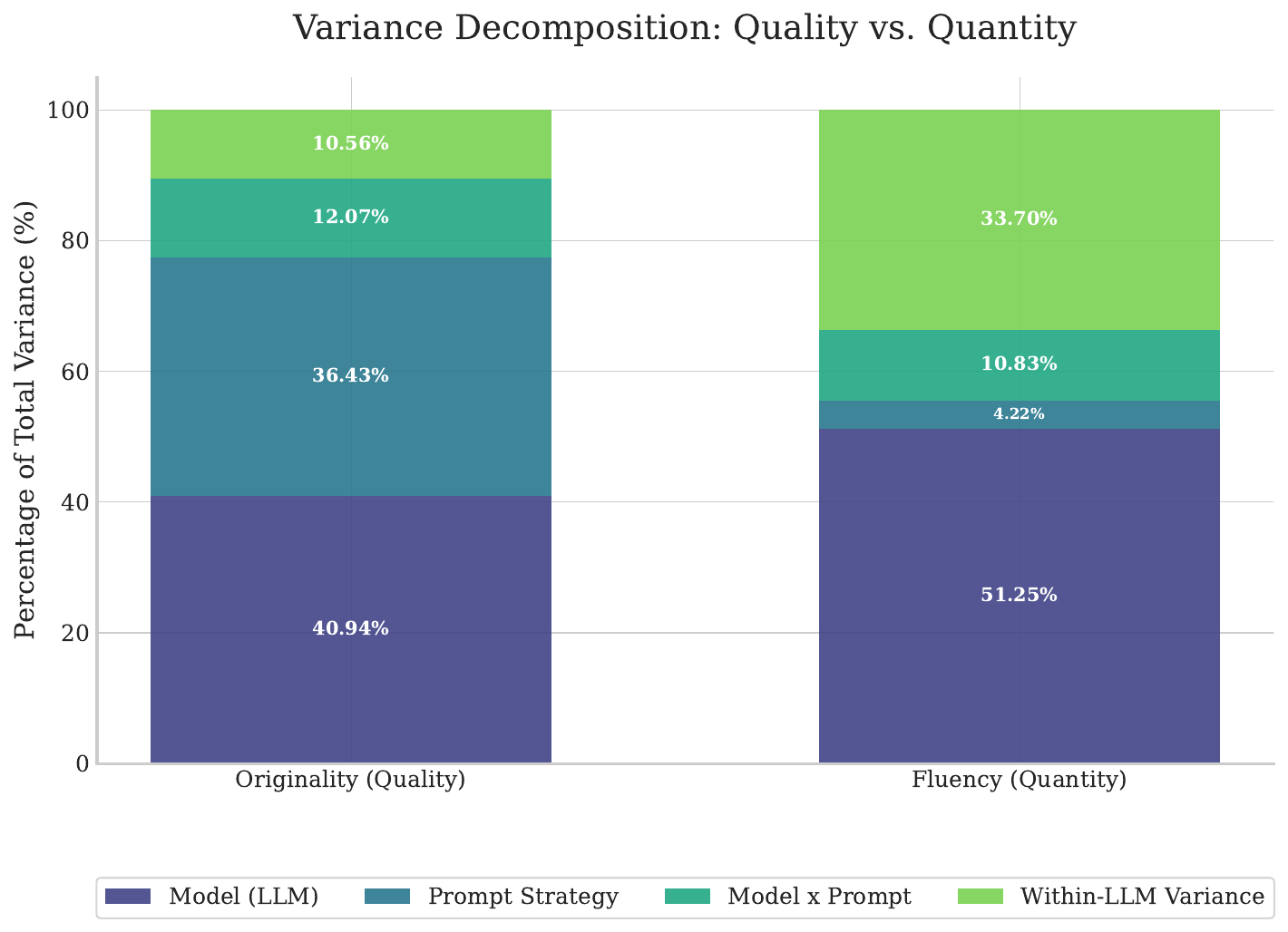}
    \caption{Variance Decomposition (ICC) for Originality and Fluency.}
    \label{fig:variance_partition}
\end{figure}

Our variance decomposition (see Figure~\ref{fig:variance_partition}) reveals a divergence between \textit{originality} and \textit{fluency}. To test for significant differences between prompts and LLMs, we fitted a linear mixed-effects model with prompt and LLM as fixed effects, including their interaction term. This analysis revealed that the means between prompts and LLMs indeed differed: For originality, we found a significant main effect of prompt, $F(9, 11870) = 4001.01, p < .001$, a significant main effect of LLM, $F(11, 11870) = 3606.72, p < .001$, and a significant interaction, $F(99, 11870) = 121.22, p < .001$. For fluency, we found a significant main effect of prompt, $F(9, 11975) = 165.74, p<.001$, a significant main effect of LLM, $F(11, 11975) = 1652.25, p<.001$, and a significant prompt × LLM interaction, $F(99, 11975)=38.80, p<.001$. 

For \textbf{Originality} (see Figure~\ref{fig:variance_partition}, and Table~\ref{tab: Descriptives_O} in Appendix for descriptive data), prompt strategy explains a substantial portion of the variance ($36.43\%$, $95\%$ CI: $35.82$--$37.14\%$), comparable to model choice ($40.94\%$, CI: $40.25$--$41.60\%$). Within-model variance accounts for $10.56\%$ (CI: $9.96$--$10.94\%$), with an additional $12.07\%$ (CI: $11.69$--$12.65\%$) driven by model$\times$prompt interactions. Formally, these interaction effects were computed as deviations from additive predictions: $I_{mp} = \bar{Y}_{mp} - (\bar{Y}_{m\cdot} + \bar{Y}_{\cdot p} - \bar{Y}_{\cdot\cdot})$, where positive values indicate specialized model-prompt affinities that exceed expectations based on marginal means. This suggests that \textit{some} prompts shift the model's stochastic distribution of creative outputs toward higher originality, but their efficacy is strongly model-contingent (see Figures~\ref{fig:model_orig} and~\ref{fig:originality_faceted} in Appendix). \textit{Some} models are more sensitive to certain prompts than others, and \textit{some} models have a smaller distribution of original ideas than others, making a prompt more or less successful for each model run.

In stark contrast, for \textbf{Fluency} (Figure~\ref{fig:variance_partition}, and Table~\ref{tab: Descriptives_F} in Appendix for descriptive data), prompt effects are negligible ($4.22\%$, CI: $3.96$--$4.52\%$), which is to expected as the prompt did not adress the fluency per se. Variance is dominated by model differences ($51.25\%$, CI: $47.05$--$54.73\%$) and within-LLM variance ($33.70\%$, CI: $29.29$--$38.53\%$). This implies that while our prompts steered for \textit{better} ideas, they have little impact on \textit{how many} ideas a model generates; volume is largely an intrinsic property of the model's decoding behavior and alisgnment (see Figures~\ref{fig:model_flu},~\ref{fig:fluency_varying}, and~\ref{fig:fluency_same} in Appendix). Our repetition analysis further clarifies this: low-fluency models often suffer from mode collapse (e.g., Gemma 3 27B repeated specific phrases in $64\%$ of outputs), whereas high-fluency models like GPT 5.1 maintained $99.5\%$ string uniqueness, suggesting that ``volume'' is a proxy for effective distribution exploration.

%\begin{table}[h]
%\centering
%\caption{Variance Decomposition (ICC) for Originality and Fluency}
%\label{tab:icc_comparison}
%\begin{tabular}{p{4.1cm}P{1.4cm}P{1.5cm}}
%\toprule
%\textbf{Source of Variance} & \textbf{Originality} & \textbf{Fluency} \\
%\midrule
%Model (LLM) & 41\% & 51\% \\
%Prompt Strategy & 36\% & 4\% \\
%Model $\times$ Prompt Interaction & 12\% & 11\% \\
%Sampling Variance (Residual) & 11\% & 34\% \\
%\bottomrule
%\end{tabular}
%\end{table}

\subsection{Mechanisms of Model Variance}
\label{sec:results:model_mechanisms}

\begin{figure}[h]
    \centering
    \includegraphics[width=\columnwidth]{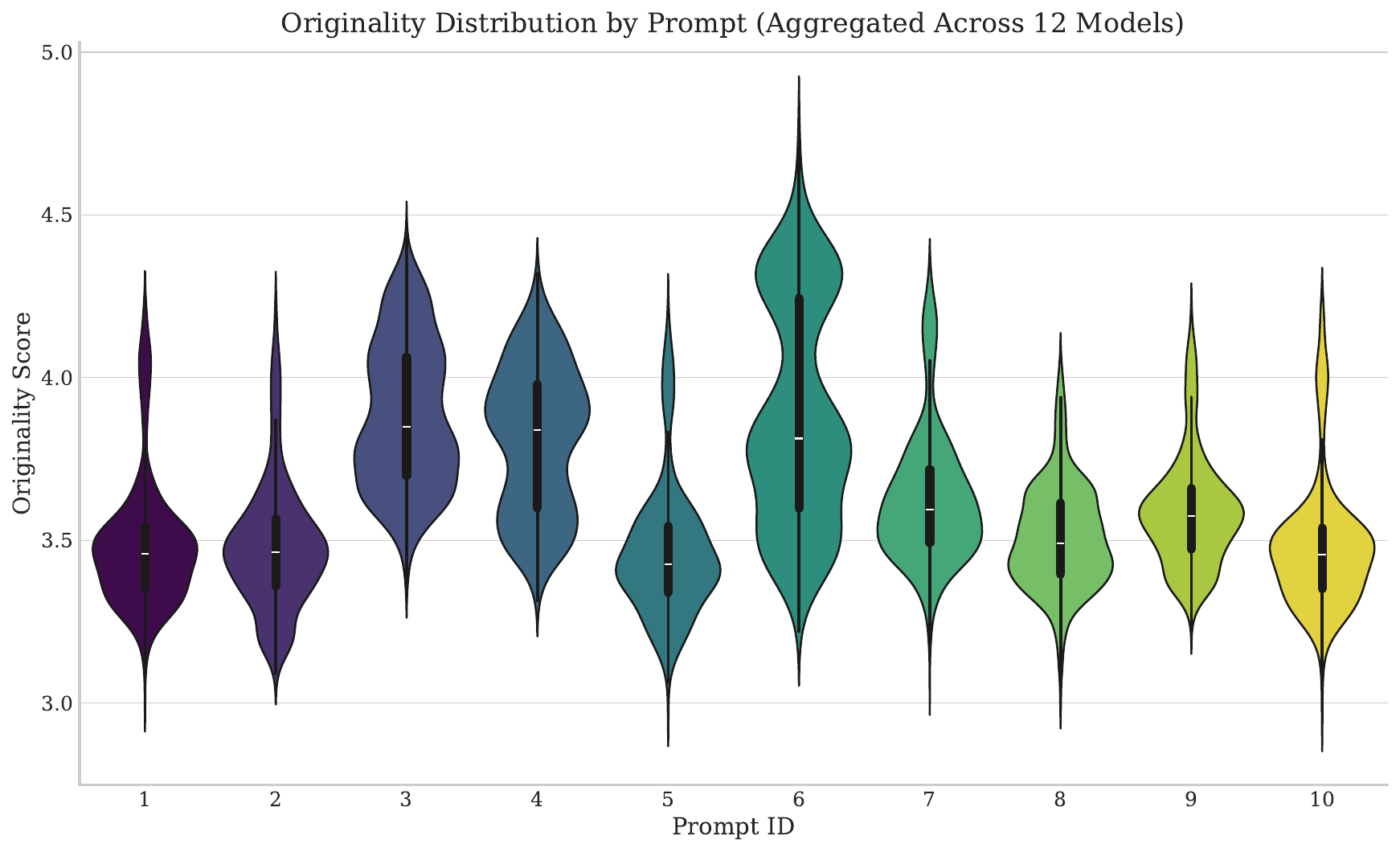}
    \caption{Violinplots displaying originality scores as a function of prompt. \textbf{Prompt Key}: \textbf{P1}: Direct Baseline (Think of uses...). \textbf{P2}: One-Shot Example (Example: Use for yarn storage...). \textbf{P3}: Heuristic/Domain (Think across domains: art, survival...). \textbf{P4}: Anticipatory (Avoid generic ideas...). \textbf{P5}: Chain-of-Thought (Think step-by-step...). \textbf{P6}: Creative Persona (You are the most creative person...). \textbf{P7}: Phrasing Variation (Synonymous rewording). \textbf{P8}: Format Constraint (No titles or colons...). \textbf{P9}: Info Order (Emphasize inventive...). \textbf{P10}: Typo Robustness (Injected noise).}
    \label{fig:prompt_originality_by_prompt_violin}
\end{figure}

The large between-model variance ($51.25\%$ for Fluency, $40.94\%$ for Originality) appears systematic rather than random, likely reflecting distinct alignment strategies (Figure~\ref{fig:quality_quantity} in Appendix). We observe several distinct model behaviors:

\paragraph{Reasoning-Oriented Architectures (High Quality, Low Fluency).} Models like Gemini 3 Pro and Qwen 3 235B Thinking prioritize depth over breadth. Gemini 3 Pro achieves the highest mean originality ($4.06$) but generates the fewest ideas (avg. $13.7$). The ``Thinking'' variant of Qwen 3 235B improves originality by $+3.4\%$ (mean $+0.119$) over the non-thinking Qwen 3 235B across all $10$ prompts, but reduces fluency by $38\%$ (from an average of $48.3$ to $30.1$ ideas). This pattern hints that deliberative reasoning steps prune lower-quality candidates in favor of focused, creative output. However, this effect appears specific to models architecturally designed for reasoning; the simple prompt asking for \textit{chain-of-thought} reasoning (P5) did not improve originality across all models (see Figure~\ref{fig:prompt_originality_by_prompt_violin}).

\paragraph{Exceptional Performance (High-Ceiling Models).} Mean performance can obscure ``high-ceiling'' capabilities. Analysis of the top $1\%$ of responses by originality ($\ge 4.36$) reveals that Grok 4.1 produced $66$ exceptional outputs ($5.5\times$ more than expected by chance) despite ranking $3$rd in overall mean. As this peak-performance just occured some of the instance runs of Grok 4.1, the average performance masks these peaks. Gemini 3 Pro also showed high-ceiling capability with $35$ exceptional responses (see Figure~\ref{fig:outliers_prompt6} in Appendix).

\paragraph{High-Fluency Architectures (High Fluency, Moderate Quality).} Models like GPT 5.1 and GPT 5.2 prioritize volume, generating $>150$ ideas per prompt on average (see Figures~\ref{fig:model_flu} and~\ref{fig:fluency_same} in Appendix). While their peak quality is high, their average score is diluted by the sheer volume of output.

\paragraph{Ranking Consistency.} Models differ fundamentally in their reliability across prompts (see Figure~\ref{fig:ranking_consistency}). Gemini 3 Pro exhibits exceptional stability (Rank Variance $= 0.18$), consistently ranking $1$st or $2$nd regardless of the prompt. In contrast, Grok 4.1 is highly volatile (Rank Variance $= 8.32$), ranking $1$st on the Persona prompt but falling to $11$th on others. This indicates that some models are ``generalist'' creators while others are ``specialists'' that require specific triggers (e.g., Persona prompt).

\paragraph{Verbosity and Metric Robustness.} Text analysis reveals that high-quality models are significantly more verbose. We find a positive correlation between idea length and originality score ($r \approx 0.1$--$0.4$) across all models, where $r$ represents Pearson's correlation coefficient measuring the linear relationship between character count and originality ratings. Gemini 3 Pro produces the longest ideas (avg. $125$ characters), while lower-performing models tend to be more concise.
To control for verbosity bias, we computed a length-adjusted originality score by regressing raw scores on character count.
Model rankings proved remarkably robust: Gemini 3 Pro retained its $\#1$ position, while Qwen 3 235B Thinking improved from $\#6$ to $\#3$, identifying it as the most concise yet original engine. This suggests that superior performance in frontier models is primarily driven by actual originality of the individual ideas, rather than mere elaboration.

\subsection{Mechanisms of Prompt Control}
\label{sec:results:prompt_mechanisms}

\begin{figure*}[ht]
    \centering
    \includegraphics[width=0.95\textwidth]{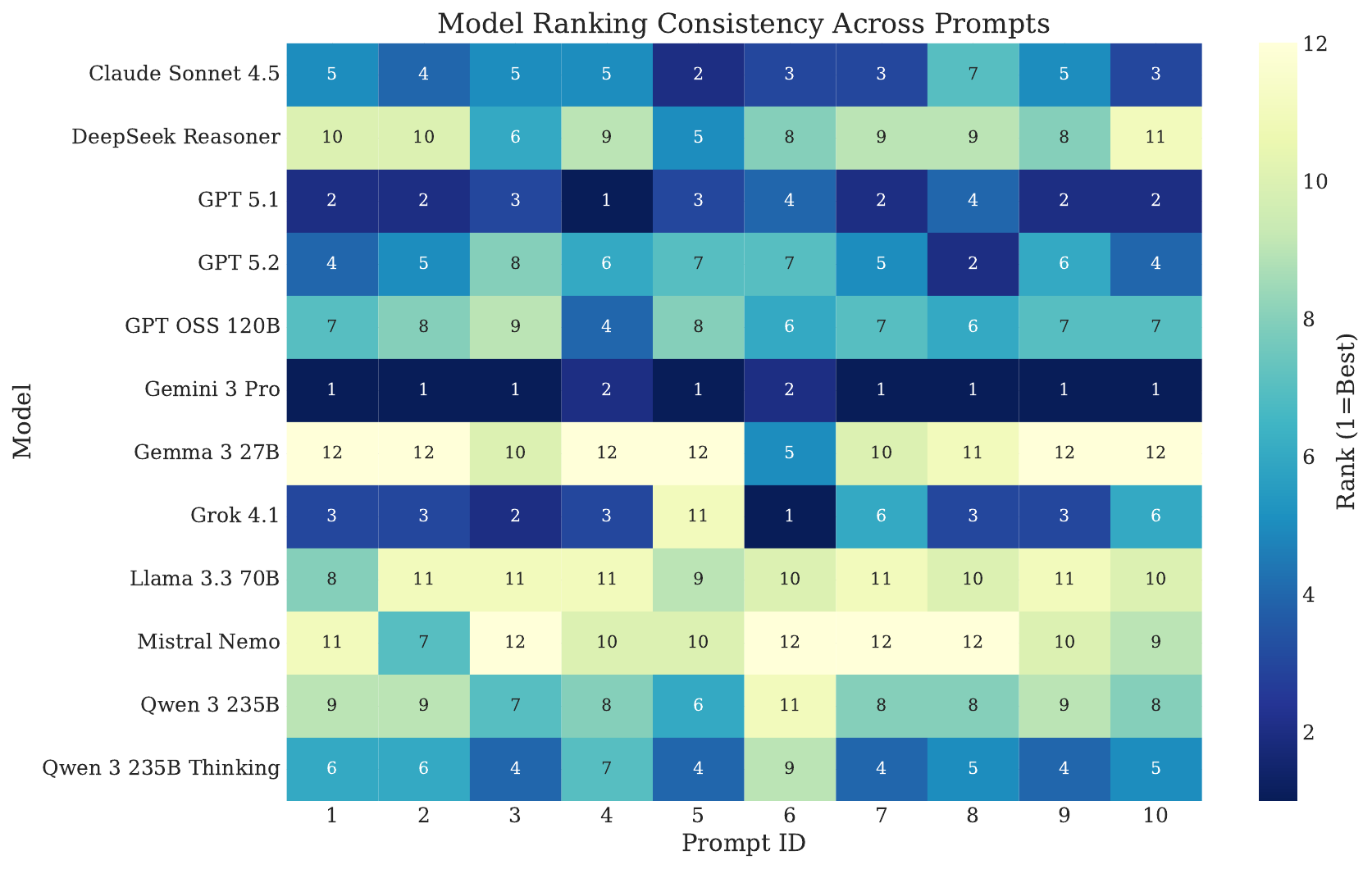}
    \caption{Heatmap of Mean Originality Scores (Model $\times$ Prompt). Prompt key as in Figure~\ref{fig:prompt_originality_by_prompt_violin}.}
    \label{fig:ranking_consistency}
\end{figure*}

The $36.43\%$ variance explained by prompts is driven by two opposing mechanisms: eliciting latent capabilities in specific models versus constraining the output space. Across all prompts, we find that strategic prompts like the Persona prompt (P6), Heuristic/Domain prompt (P3) and the Anticipatory prompt (P4) elicited higher level of originality, but also higher between-model variance than the baseline. Instructions with one example (P2) or the prompt to think step-by-step (P5) did not improve the output quality significantly. Similarly, the minor prompting variations are on-par with the baseline, suggesting that they do not have a significant effect on the output quality (see Figures~\ref{fig:prompt_originality_by_prompt_violin} and~\ref{fig:ranking_consistency}).

\paragraph{Discriminative Capability Elicitation: Prompt 6 (Persona).} 
The Persona prompt (P6: ``You are the most creative person...'') can be interpreted as a discriminative stress test. It yields the highest between-model variance ($0.112$ vs. baseline P1 variance of $0.037$). Capable models like Grok 4.1 and Claude Sonnet 4.5 respond with massive effect sizes (Cohen's $d > 2.0$, calculated as the standardized difference between the model's P6 performance and its overall mean), producing their most-original work (see Figure~\ref{fig:outliers_prompt6} in Appendix). Specifically, Gemma 3 27B shows a jump ($d=2.11$), effectively ``waking up'' under this prompt (see Figure~\ref{fig:ranking_consistency} in Appendix). In contrast, smaller models like Llama 3.3 70B show negligible improvement ($d=0.48$). This explains the significant Model$\times$Prompt interaction (see Figure~\ref{fig:ranking_consistency}): ``Persona'' prompts (P6) do not universally improve creativity; they reveal latent creative capacity only in models capable of role-play.

\paragraph{Structural Constraint: Prompt 8 (Format).}
Conversely, the Format Constraint prompt (P8: ``no titles or colons'') systematically degrades fluency (but not originality, see Figure~\ref{fig:prompt8} in Appendix). The mechanism is structural collapse: compliance with the ``no colons'' instruction was near-perfect ($99.99\%$ of ideas contained no colons), forcing models to abandon their diverse default formats. This led to a convergence on identical phrasings, with over $100$ exact duplicates of common ideas (e.g., ``Build a miniature greenhouse...''). By restricting the syntactic search space, P8 inadvertently restricted the semantic search space, reducing idea uniqueness by $5.9$ percentage points relative to the baseline (P1). This is important to note, as this was not intended by the prompt, but rather a byproduct of the model's default formatting behavior. Especially those models that are more verbose and usually show a high-colon usage, like Gemini 3 Pro, are more negatively affected by this prompt.

\section{Discussion}
\label{sec:discussion}

The field of prompt engineering has largely operated on the premise that prompts act as close-to-deterministic programs: find the right instruction, and the model will reliably execute it. Our variance decomposition challenges this view. While prompts do explain $36.43\%$ of variance in originality, this effect is accompanied by substantial model differences ($40.94$--$51.25\%$) and within-LLM variance ($10.56$--$33.70\%$). This suggests that prompting is less about programming a machine and more about steering a probabilistic distribution.

\paragraph{Probabilistic Prompting: Shaping Distributions, Not Outputs.}
Our findings show that a prompt defines a \textit{distribution} of likely outputs, not a single result. For discriminative prompts like the ``Creative Persona'' (P6), this distribution shifts upward in quality but also expands in variance. This means that ``better'' prompts often yield \textit{less} predictable behavior. This creates a paradox for developers: the prompts that unlock the highest creativity (P6) are also the least stable, while the prompts that aim at consistency (P8) actively suppress idea-diversity. Effective prompting systems must therefore treat outputs as samples, with multiple candidates ($N>1$) generated to capture the full shape of the distribution and allow for statistical inference.

\paragraph{The Need for Discriminative Benchmarking.}
Current benchmarks often use generic instructions that fail to separate model capabilities. We found that baseline prompts (P1) yielded low between-model variance ($0.037$), making them poor tools for comparison. In contrast, the ``Persona'' prompt (P6) acted as a stress test (variance $0.112$), revealing substantial differences in how models respond to identity-based framing. While we cannot definitively attribute Gemini and Grok's superior performance on persona prompts to specific training procedures, their responsiveness suggests these models may possess stronger capabilities in contextual self-modification or instruction interpretation when framed through identity rather than direct commands. The Persona prompt appears to test whether models can leverage identity cues to access different behavioral modes, distinguishing models that treat personas as meaningful context from those that process them as superficial instruction variants. Future creativity benchmarks should therefore prioritize \textit{discriminative items}, as prompts that reveal latent capability differences through varied framing approaches.

\paragraph{Beyond $N=1$: Guidelines for Sample Complexity.}
This has direct implications for reproducibility: if within-LLM variance explains $10.56$--$33.70\%$ of total variance, many reported prompt or model comparisons in the literature may reflect sampling noise rather than genuine effects. Current evaluation practices that rely on single generations ($N=1$) effectively treat LLM outputs as deterministic point estimates. Our results demonstrate that this approach is statistically unreliable, particularly for creative tasks. However, the risk of $N=1$ evaluation is not uniform across architectures. For ``stable generalists'' like Gemini 3 Pro (Rank Variance $= 0.18$), small sample sizes may suffice for coarse ranking. In contrast, for ``high-variance specialists'' like Grok 4.1 (Rank Variance $= 8.32$), a single sample carries a high risk of rank-reversal, potentially mischaracterizing a top-tier model as average or vice versa. We find that while mean originality stabilizes relatively quickly, capturing the true width of the output distribution and the probability of exceptional responses (top $1\%$) requires larger sample sizes ($N \ge 50$). We recommend that future benchmarks report not only mean scores but also the variance components and associated confidence intervals to allow for valid cross-model inference.

\paragraph{Limitations.}

\textit{Task scope.} Our study focused on a single creative domain (alternative uses for plastic bottles) to enable deep repeated sampling ($N=12{,}000$). While prior work suggests that AUT item selection has minimal effect on creativity scores \citep{organisciakOpenCreativityScoring2025}, future studies should validate these findings across multiple AUT items and creative domains. The fundamental principle, that within-LLM variance competes with prompt control, likely generalizes to all open-ended generation, but the specific ICC values may shift for coding or math tasks. Further, we did not directly prompt for fluency, so this measures default output volume rather than capacity to generate more ideas on demand.

\textit{Confounded variance components.} Our ``model'' factor bundles architecture, training data, RLHF/alignment, default parameters, system prompts, and serving infrastructure. We retain ``model'' as it reflects common usage, though variance cannot be attributed to architecture alone. Our ``within-LLM variance'' is technically variance within a model--prompt condition, emphasizing the inherent sampling stochasticity. However, this component conflates true sampling noise (temperature, nucleus sampling) with unmodeled factors (e.g., server load, API versioning). Disentangling architecture effects requires controlled experiments with matched parameters, left for future work.

\textit{Statistical model.} Our fluency analysis uses a linear mixed model on count data. While ICC decomposition remains valid for comparative purposes, a generalized linear mixed model (GLMM) with Poisson or negative binomial distribution would more appropriately model the discrete, non-negative nature of idea counts. Sensitivity analyses suggest our qualitative conclusions are robust, but future work should employ distributional assumptions even better matched to the outcome variable.

\section{Conclusion}
\label{sec:conclusion}

We provide a large-scale quantification of how variance in LLM creative output partitions across prompt, model, and within-LLM sources. Our results demystify the ``magic'' of prompt engineering: prompts are powerful but bounded levers. They can effectively steer originality ($36.43\%$ variance explained) but output is also heavily influenced by model choice ($40.94$--$51.25\%$) and within-LLM variance ($10.56$--$33.70\%$). Building on existing prompt-based approaches, explicitly accounting for the probabilistic nature of these systems and examining the variance they produce supports more robust, statistically grounded design practices.

% Impact Statement is required
\section*{Impact Statement}

This work demonstrates that variability in large language model outputs, especially in open-ended creative tasks, arises from three sources: (1) prompt strategy, (2) model choice, and (3) within-LLM stochasticity through sampling variance. By quantifying within-LLM variance alongside prompt and model effects, the study challenges evaluation practices that rely on single-sample decoding. The practical implication is simple: if within-LLM variance accounts for $10$--$34\%$ of output variability, users dissatisfied with an LLM response might need not revise their prompt directly, instead simply regenerating with the same prompt may yield substantially different, and potentially better, results.

% Uncomment the following section for camera-ready version:

% \section*{Acknowledgements}
% This work was partially supported by the Federal Ministry of Research, Technology and Space under the grant 16DII133, founded by the German Federal Ministry of Education and Research in 2017, by Deutsche Forschungsgemeinschaft under grants 496119880 (VisualMine), 531115272 (ProImpact), and SFB 1404/2 (FONDA), as well as Deutsche Forschungsgemeinschaft (DFG) through the DFG Cluster of Excellence MATH+ (grant number EXC-2046/1, project ID 390685689), as well as the Zuse Institute Berlin via the RISE@ZIB services hosting the LLM models.

% Bibliography
\bibliography{Prompting}
\bibliographystyle{icml2026}

%%%%%%%%%%%%%%%%%%%%%%%%%%%%%%%%%%%%%%%%%%%%%%%%%%%%%%%%%%%%%%%%%%%%%%%%%%%%%%%
%%%%%%%%%%%%%%%%%%%%%%%%%%%%%%%%%%%%%%%%%%%%%%%%%%%%%%%%%%%%%%%%%%%%%%%%%%%%%%%
% APPENDIX
%%%%%%%%%%%%%%%%%%%%%%%%%%%%%%%%%%%%%%%%%%%%%%%%%%%%%%%%%%%%%%%%%%%%%%%%%%%%%%%
%%%%%%%%%%%%%%%%%%%%%%%%%%%%%%%%%%%%%%%%%%%%%%%%%%%%%%%%%%%%%%%%%%%%%%%%%%%%%%%
\newpage
\appendix
\onecolumn
\section{Appendix}
\label{app:appendix}

\subsection{Full Prompt Texts}
\label{app:full-prompts}

\begin{table*}[ht]
    \centering
    \caption{Core Prompting Strategies (P1--P6) -- Full Text}
    \label{tab:prompting-strategies-full}
    \begin{tabular}{lp{0.09\linewidth}p{0.26\linewidth}p{0.54\linewidth}}
    \toprule
    \textbf{ID} & \textbf{Strategy} & \textbf{Category Logic Description} & \textbf{Prompt Example (abridged)} \\
    \midrule
    
    P1 & Direct \newline Instruction & 
    Baseline: Gives a clear task with explicit output format; relies on default LLM behavior. &
    \footnotesize\texttt{Think of all kinds of things you could do with a plastic bottle. ...$^1$} \\
    \addlinespace
    
    P2 & One-Shot \newline Example & 
    Example-Based: Includes a single concrete example to anchor the model's response style. &
    \footnotesize\texttt{Example – A plastic bottle can be used to store and dispense yarn or string without tangling.  ...$^1$} \\
    \addlinespace
    
    P3 & Heuristic \newline Prompt & 
    Domain Exploration: Provides mental heuristics and creativity-stimulating cues (e.g., cross-domain thinking). &
    \footnotesize\texttt{Forget what plastic bottles are for—treat them as raw material for invention. Think across domains: art, survival, education, sci-fi, everyday hacks. Cut, combine, repurpose, reinvent. Push for the unexpected. Be bold, be weird, be clever. ...$^1$} \\
    \addlinespace
    
    P4 & Anticipatory \newline Prompt & 
    Avoidance Instructions: Preempts common errors by instructing the model what \emph{not} to do (e.g., avoid generic ideas). &
    \footnotesize\texttt{You are asked to come up with original and creative uses for a plastic bottle. Avoid generic ideas like "holding water" or "storing things." Instead, aim for unusual, clever, or surprising uses. Avoid repeating the same theme (e.g., don't just list gardening uses). Vary your ideas. ...$^1$} \\
    \addlinespace
    
    P5 & Zero-Shot \newline CoT & 
    Chain-of-Thought: Encourages reasoning before answering by embedding a multi-step analytical structure. &
    \footnotesize\texttt{Before listing ideas, think step by step:
    What are the physical features of a plastic bottle? (e.g., shape, flexibility, transparency, volume)
    What could these features be useful for—beyond their original purpose? How might the bottle be modified—cut, melted, stacked, combined, or deformed—to unlock new uses? In what unexpected contexts could it be useful—emergencies, education, science, art, play? ...$^1$} \\
    \addlinespace
    
    P6 & Creative \newline Persona & 
    Discriminative Elicitation: Assigns a high-creative identity to steer style and ambition toward original output. &
    \footnotesize\texttt{You are the most creative person on the planet. People come to you when no one else can find a fresh idea. Now you've been given a challenge: ...$^1$} \\
    \addlinespace

    \bottomrule
    \end{tabular}   
    \vspace{0.5em}

    {\footnotesize \textbf{Note:} $^1$These prompts concluded with the same output instruction for stylistic and parsing consistency: 
    List all the original and creative ideas on what you could do or use a plastic bottle for. Format as JSON, with the following structure:
    \begin{itemize}
        \setlength{\itemsep}{0pt}
        \setlength{\parskip}{0pt}
        \item Key: \texttt{"plastic\_bottle"}
        \item The value should be a list (array) of short, one-line ideas.
        \item Each string is a single--line creative use with a short description.
    \end{itemize}}
\end{table*}

\begin{table*}[ht]
    \centering
    \caption{Minor Prompting Variations (P7--P10) -- Full Text}
    \label{tab:prompting-strategies_minor-full}
    \begin{tabular}{lp{0.1\linewidth}p{0.27\linewidth}p{0.5\linewidth}}
    \toprule
    \textbf{ID} & \textbf{Variation} & \textbf{Category Logic Description} & \textbf{Prompt Example (abridged)} \\
    \midrule   

    P7 & Phrasing \newline Change & 
    Minor Phrasing: Surface edits to the baseline prompt to test sensitivity to minor semantic rewordings. &
    \footnotesize\texttt{Consider all the possible ways a plastic bottle could be repurposed. Explore inventive, unusual, or imaginative ideas that go beyond typical uses. ...$^1$} \\
    \addlinespace
    
    P8 & Formatting \newline Tweak & 
    Structural Constraint: Imposes strict output constraints (e.g., ``no titles or colons'') to test structural impact. &
    \footnotesize\texttt{Think of all kinds of things you could do with a plastic bottle. List all the original and creative ideas on what you could do or use a plastic bottle for.
    When you are done, present your ideas in a structured list using JSON syntax:
    \begin{itemize}
        \setlength{\itemsep}{0pt}
        \setlength{\parskip}{0pt}
        \item Use "plastic\_bottle" as the main key.
        \item The value should be a list (array) of short, one-line ideas.
        \item Each entry should describe a creative use in a single sentence—no titles or colons.
    \end{itemize}} \\
    \addlinespace
    
    P9 & Information \newline Order & 
    Information order: Reorders elements but keep the same information. &
    \footnotesize\texttt{List all the original and creative ideas you can think of for how a plastic bottle could be used. Focus on inventive, surprising, or resourceful purposes. ...$^1$...
    Think broadly and imaginatively about all the things one could do with a plastic bottle. } \\
    \addlinespace
    
    P10 & Random \newline Errors & 
    Typo-robustness: Injects typographical and syntactic noise to assess model error-tolerance. &
    \footnotesize\texttt{Thnik of ; all kindss of thing,s you could do : with a plastic bottle.
    List all the orginal ,, and creativ ideas on what you could do or use a plastic bottel for.
    Frormat as JSON ;
        \begin{itemize}
            \setlength{\itemsep}{0pt}
            \setlength{\parskip}{0pt}
            \item Key ::  "plastic\_bottle"
            \item Valuue ,: array of stringz
            \item Each  string is a single--line ; creative use with a shrt descripshun.
        \end{itemize}} \\
    \bottomrule
    \end{tabular}   
    \vspace{0.5em}

    \textbf{Note:} $^1$Same as Table~\ref{tab:prompting-strategies-full}.
\end{table*}

\clearpage
\subsection{Originality and Fluency Distributions}

\begin{table}[ht]
\caption{Originality Scores by LLM and Prompt: Mean (SD)}
\centering
\footnotesize
\begin{tabular}[t]{l@{\hspace{2pt}}p{0.12\linewidth}@{\hspace{2pt}}c@{\hspace{2pt}}c@{\hspace{2pt}}c@{\hspace{2pt}}c@{\hspace{2pt}}c@{\hspace{2pt}}c@{\hspace{2pt}}c@{\hspace{2pt}}c@{\hspace{2pt}}c@{\hspace{2pt}}c@{\hspace{2pt}}c}
\toprule
  &  & P1 & P2 & P3  & P4  & P5 & P6 & P7 & P8  & P9  & P10  & \\
  & Model & (Baseline) & (Example) &  (Heuristic) & (Anticipatory) &  (CoT) &  (Persona) &  (Phrasing) &  (Formatting) & (Order) &  (Errors) & Mean \\
\midrule
1 & Claude Sonnet & 3.50 & 3.49 & 4.01 & 3.91 & 3.57 & 4.26 & 3.71 & 3.48 & 3.61 & 3.51 & 3.70\\
  & 4.5       & (0.06) & (0.06) & (0.11) & (0.12) & (0.05) & (0.11) & (0.08) & (0.07) & (0.07) & (0.06) & (0.27)\\
2 & DeepSeek  & 3.38 & 3.40 & 3.88 & 3.62 & 3.44 & 3.73 & 3.53 & 3.42 & 3.55 & 3.35 & 3.53\\
  &        Reasoner V3.2        & (0.09) & (0.09) & (0.14) & (0.12) & (0.08) & (0.19) & (0.11) & (0.12) & (0.10) & (0.09) & (0.20)\\
3 & Gemini 3 Pro & 4.03 & 3.95 & 4.24 & 4.09 & 3.98 & 4.34 & 4.14 & 3.82 & 3.99 & 4.00 & 4.06\\
  &         & (0.10) & (0.12) & (0.07) & (0.10) & (0.10) & (0.08) & (0.09) & (0.13) & (0.09) & (0.12) & (0.17)\\
4 & Gemma 3 27B & 3.30 & 3.19 & 3.66 & 3.49 & 3.22 & 3.84 & 3.47 & 3.37 & 3.37 & 3.31 & 3.42\\
  &        & (0.04) & (0.03) & (0.06) & (0.05) & (0.05) & (0.07) & (0.04) & (0.05) & (0.03) & (0.06) & (0.20)\\
5 & GPT 5.1 & 3.62 & 3.66 & 4.06 & 4.11 & 3.54 & 4.25 & 3.80 & 3.57 & 3.71 & 3.58 & 3.81\\
  &          & (0.06) & (0.06) & (0.08) & (0.09) & (0.07) & (0.09) & (0.08) & (0.07) & (0.06) & (0.07) & (0.26)\\
6 & GPT 5.2 & 3.50 & 3.48 & 3.79 & 3.88 & 3.42 & 3.78 & 3.66 & 3.65 & 3.60 & 3.50 & 3.63\\
  &          & (0.03) & (0.04) & (0.05) & (0.05) & (0.04) & (0.08) & (0.04) & (0.03) & (0.03) & (0.04) & (0.15)\\
7 & GPT OSS 120B & 3.42 & 3.42 & 3.75 & 3.97 & 3.39 & 3.83 & 3.56 & 3.48 & 3.55 & 3.46 & 3.58\\
  &          & (0.07) & (0.07) & (0.10) & (0.10) & (0.09) & (0.11) & (0.07) & (0.07) & (0.07) & (0.09) & (0.21)\\
8 & Grok 4.1 & 3.52 & 3.56 & 4.11 & 3.97 & 3.30 & 4.39 & 3.58 & 3.65 & 3.64 & 3.49 & 3.72\\
  &      & (0.09) & (0.07) & (0.14) & (0.09) & (0.09) & (0.15) & (0.10) & (0.06) & (0.09) & (0.07) & (0.33)\\
9 & Llama 3.3 70B & 3.39 & 3.34 & 3.65 & 3.56 & 3.38 & 3.51 & 3.47 & 3.40 & 3.44 & 3.35 & 3.45\\
  &       & (0.07) & (0.06) & (0.10) & (0.09) & (0.06) & (0.14) & (0.07) & (0.07) & (0.07) & (0.07) & (0.13)\\
10 & Mistral Nemo & 3.33 & 3.42 & 3.61 & 3.57 & 3.35 & 3.50 & 3.41 & 3.31 & 3.49 & 3.38 & 3.44\\
   &         & (0.10) & (0.14) & (0.10) & (0.11) & (0.15) & (0.13) & (0.11) & (0.11) & (0.10) & (0.15) & (0.15)\\
11 & Qwen 3 235B & 3.39 & 3.40 & 3.82 & 3.73 & 3.43 & 3.50 & 3.55 & 3.45 & 3.53 & 3.39 & 3.52\\
   &       & (0.11) & (0.11) & (0.13) & (0.10) & (0.11) & (0.08) & (0.10) & (0.06) & (0.09) & (0.16) & (0.18)\\
12 & Qwen 3 235B & 3.48 & 3.47 & 4.03 & 3.81 & 3.53 & 3.72 & 3.68 & 3.52 & 3.64 & 3.50 & 3.64\\
   & Thinking             & (0.10) & (0.11) & (0.17) & (0.13) & (0.12) & (0.17) & (0.12) & (0.11) & (0.13) & (0.10) & (0.21)\\
\midrule
 & Prompt Mean & 3.49 & 3.48 & 3.88 & 3.81 & 3.46 & 3.89 & 3.63 & 3.51 & 3.59 & 3.48 & \\
 &             & (0.20) & (0.20) & (0.22) & (0.22) & (0.20) & (0.34) & (0.21) & (0.16) & (0.17) & (0.20) & \\
\bottomrule
\end{tabular}
\label{tab: Descriptives_O}
\end{table}

\begin{table}[ht]
\caption{Fluency Scores by LLM and Prompt: Mean (SD)}
\centering
\footnotesize
\begin{tabular}[t]{l@{\hspace{2pt}}p{0.12\linewidth}@{\hspace{2pt}}c@{\hspace{2pt}}c@{\hspace{2pt}}c@{\hspace{2pt}}c@{\hspace{2pt}}c@{\hspace{2pt}}c@{\hspace{2pt}}c@{\hspace{2pt}}c@{\hspace{2pt}}c@{\hspace{2pt}}c@{\hspace{2pt}}c}
\toprule
  &  & P1 & P2 & P3  & P4  & P5 & P6 & P7 & P8  & P9  & P10  & \\
  & Model & (Baseline) & (Example) &  (Heuristic) & (Anticipatory) &  (CoT) &  (Persona) &  (Phrasing) &  (Formatting) & (Order) &  (Errors) & Mean \\
\midrule
1 & Claude Sonnet & 44.90 & 39.35 & 59.41 & 18.56 & 76.25 & 14.98 & 51.57 & 51.72 & 45.44 & 34.83 & 43.70\\
  & 4.5       & (7.12) & (9.66) & (17.27) & (3.06) & (19.90) & (1.85) & (6.74) & (16.00) & (8.32) & (9.59) & (20.73)\\
2 & DeepSeek  & 21.95 & 19.04 & 22.24 & 14.02 & 33.36 & 13.70 & 24.60 & 25.90 & 22.60 & 20.20 & 21.76\\
  &    Reasoner V3.2        & (5.12) & (4.90) & (7.92) & (4.52) & (11.33) & (3.99) & (8.32) & (6.98) & (5.98) & (3.69) & (8.58)\\
3 & Gemini 3 Pro & 13.34 & 10.72 & 15.64 & 10.16 & 15.28 & 9.93 & 15.13 & 18.05 & 17.33 & 11.49 & 13.70\\
  &         & (2.59) & (1.31) & (2.56) & (0.80) & (2.35) & (0.58) & (2.50) & (2.75) & (2.67) & (2.08) & (3.57)\\
4 & Gemma 3 27B & 31.20 & 21.76 & 47.79 & 24.60 & 36.95 & 22.37 & 36.61 & 33.43 & 27.99 & 31.63 & 31.43\\
  &        & (5.44) & (2.42) & (5.06) & (1.72) & (3.34) & (2.64) & (3.92) & (6.02) & (1.76) & (7.19) & (8.66)\\
5 & GPT 5.1 & 216.01 & 183.24 & 131.36 & 60.90 & 248.27 & 54.57 & 192.22 & 126.03 & 167.89 & 149.43 & 152.99\\
  &          & (131.65) & (154.65) & (52.38) & (34.68) & (187.79) & (29.16) & (125.48) & (59.90) & (82.12) & (92.40) & (122.48)\\
6 & GPT 5.2 & 176.58 & 168.64 & 211.19 & 64.65 & 139.86 & 42.34 & 230.23 & 142.99 & 219.32 & 179.45 & 157.53\\
  &          & (29.72) & (33.23) & (43.06) & (10.29) & (24.55) & (8.59) & (51.61) & (26.02) & (36.01) & (29.38) & (67.41)\\
7 & GPT OSS 120B & 37.02 & 31.19 & 35.75 & 19.72 & 37.61 & 14.70 & 37.66 & 32.28 & 35.74 & 25.35 & 30.70\\
  &          & (4.33) & (4.38) & (3.64) & (2.17) & (5.41) & (2.86) & (3.28) & (5.20) & (4.01) & (4.62) & (8.73)\\
8 & Grok 4.1 & 21.94 & 24.93 & 25.60 & 14.80 & 29.89 & 10.15 & 29.48 & 21.97 & 26.48 & 23.19 & 22.84\\
  &      & (3.19) & (3.89) & (5.22) & (1.89) & (4.85) & (0.86) & (6.12) & (3.46) & (5.12) & (3.46) & (7.16)\\
9 & Llama 3.3 70B & 19.24 & 17.65 & 27.94 & 18.01 & 21.02 & 14.37 & 22.56 & 19.10 & 21.59 & 25.08 & 20.66\\
  &       & (1.41) & (1.51) & (3.85) & (1.66) & (2.12) & (1.50) & (2.08) & (1.05) & (2.64) & (2.66) & (4.30)\\
10 & Mistral Nemo & 11.17 & 9.29 & 13.67 & 10.13 & 12.07 & 9.48 & 12.16 & 12.86 & 11.84 & 9.86 & 11.25\\
   &         & (1.51) & (1.13) & (2.27) & (0.94) & (2.19) & (0.95) & (1.88) & (2.72) & (1.92) & (1.01) & (2.26)\\
11 & Qwen 3 235B & 66.46 & 39.63 & 48.46 & 18.89 & 51.29 & 18.69 & 67.32 & 35.38 & 23.39 & 114.40 & 48.30\\
   &       & (63.15) & (43.88) & (45.51) & (2.98) & (41.39) & (6.87) & (101.08) & (36.41) & (12.06) & (103.40) & (62.75)\\
12 & Qwen 3 235B & 30.93 & 28.42 & 25.17 & 17.30 & 31.99 & 20.51 & 30.87 & 35.03 & 39.06 & 41.40 & 30.05\\
   & Thinking             & (26.45) & (42.43) & (15.86) & (4.01) & (38.61) & (6.86) & (32.43) & (29.03) & (136.91) & (68.55) & (54.77)\\
\midrule
 & Prompt Mean & 57.55 & 49.50 & 55.44 & 24.31 & 61.32 & 20.49 & 62.52 & 46.26 & 55.02 & 55.58 & \\
 &             & (77.66) & (75.39) & (61.21) & (20.63) & (87.09) & (16.29) & (84.58) & (47.28) & (79.34) & (71.78) & \\
\bottomrule
\end{tabular}
\label{tab: Descriptives_F}
\end{table}

\begin{figure*}[ht]
    \centering
    \includegraphics[width=0.6\textwidth]{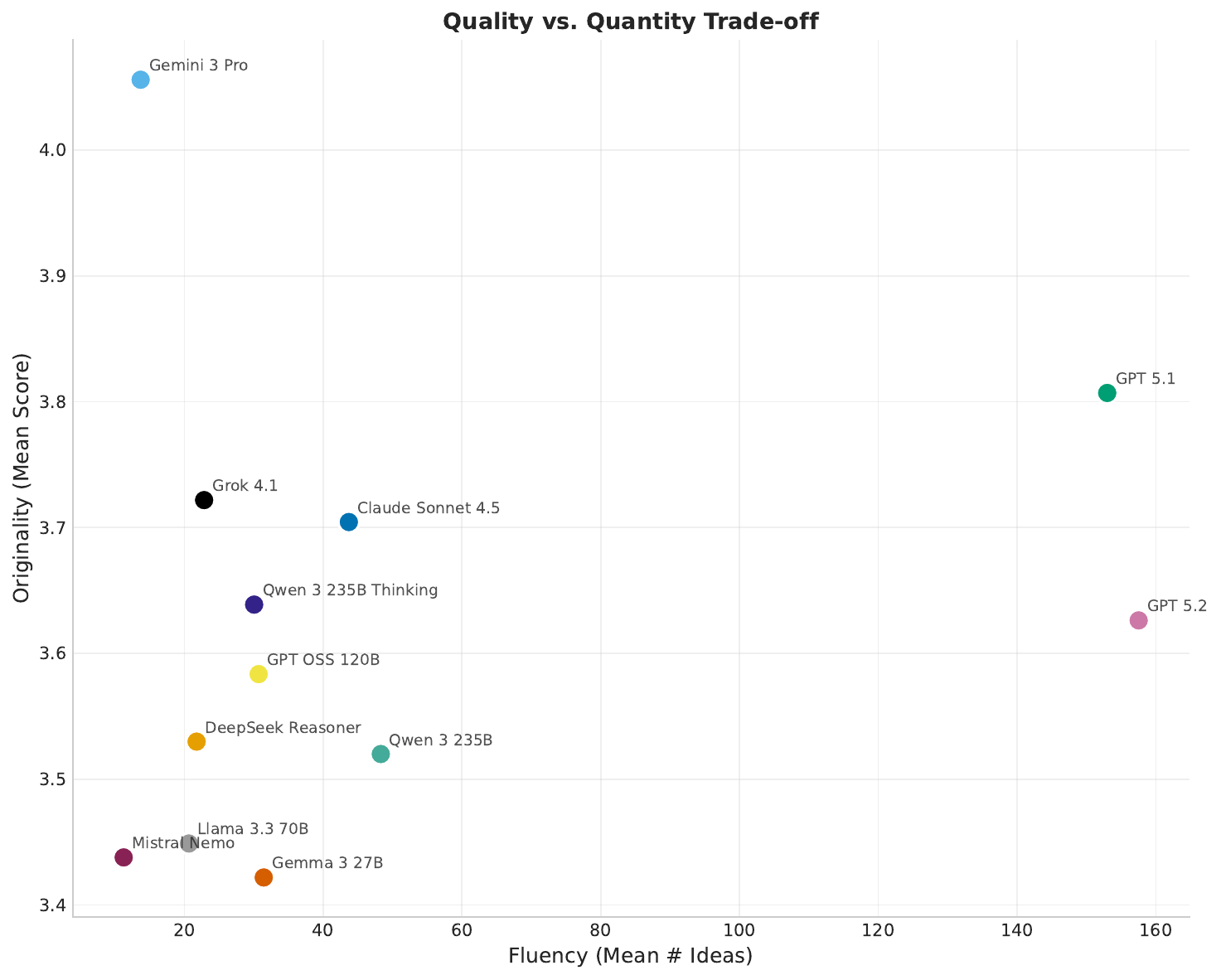}
    \caption{Quality vs. Quantity Trade-off. Models cluster into ``Reasoning-Oriented'' (top-left: high originality, low fluency) and ``High-Fluency'' (middle-right: high volume, moderate quality) architectures.}
    \label{fig:quality_quantity}
\end{figure*}

\begin{figure*}[ht]
    \centering
    \begin{minipage}{0.48\textwidth}
        \centering
        \includegraphics[width=\linewidth]{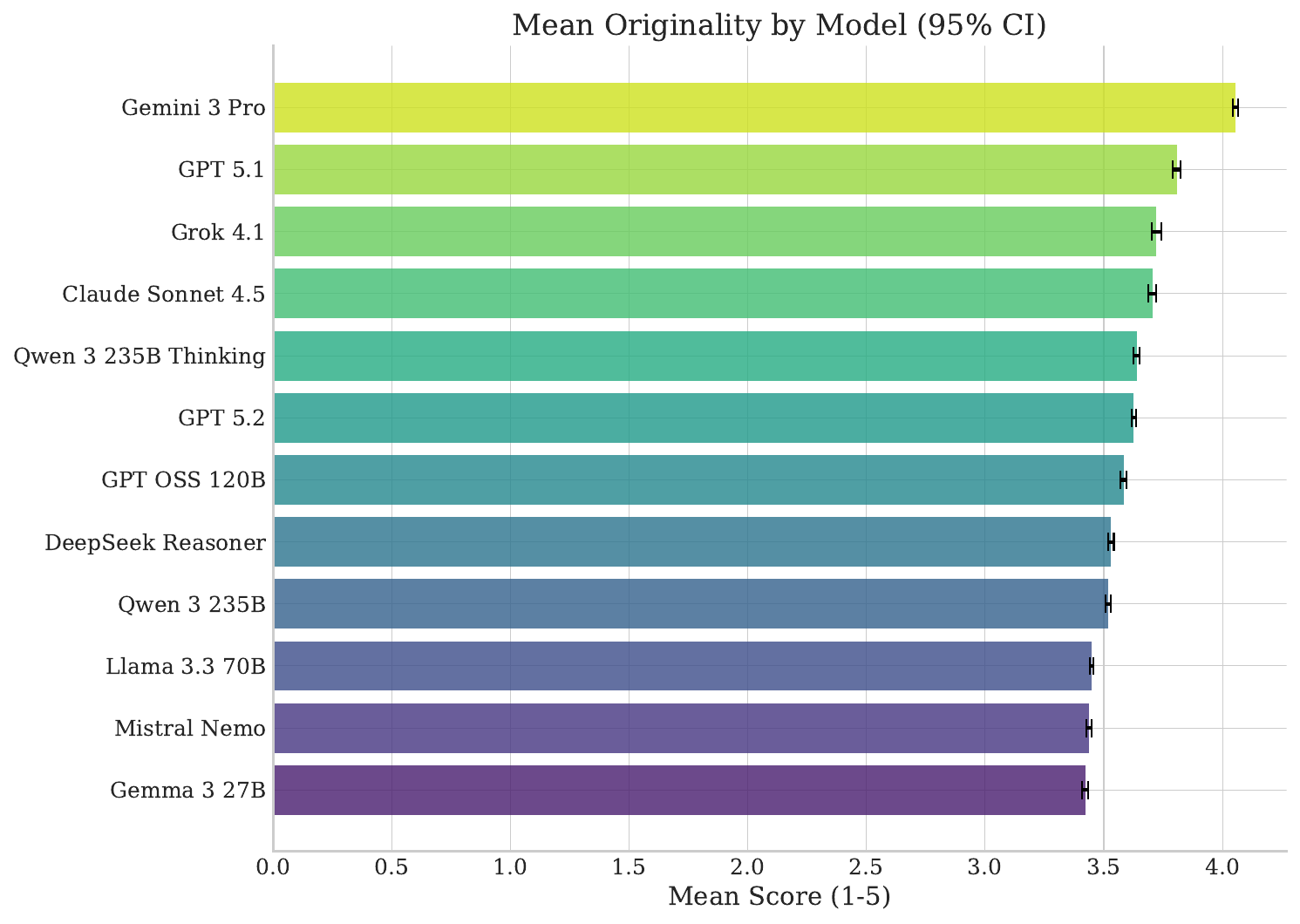}
        \caption{Mean Originality by Model (95\% CI).}
        \label{fig:model_orig}
    \end{minipage}\hfill
    \begin{minipage}{0.48\textwidth}
        \centering
        \includegraphics[width=\linewidth]{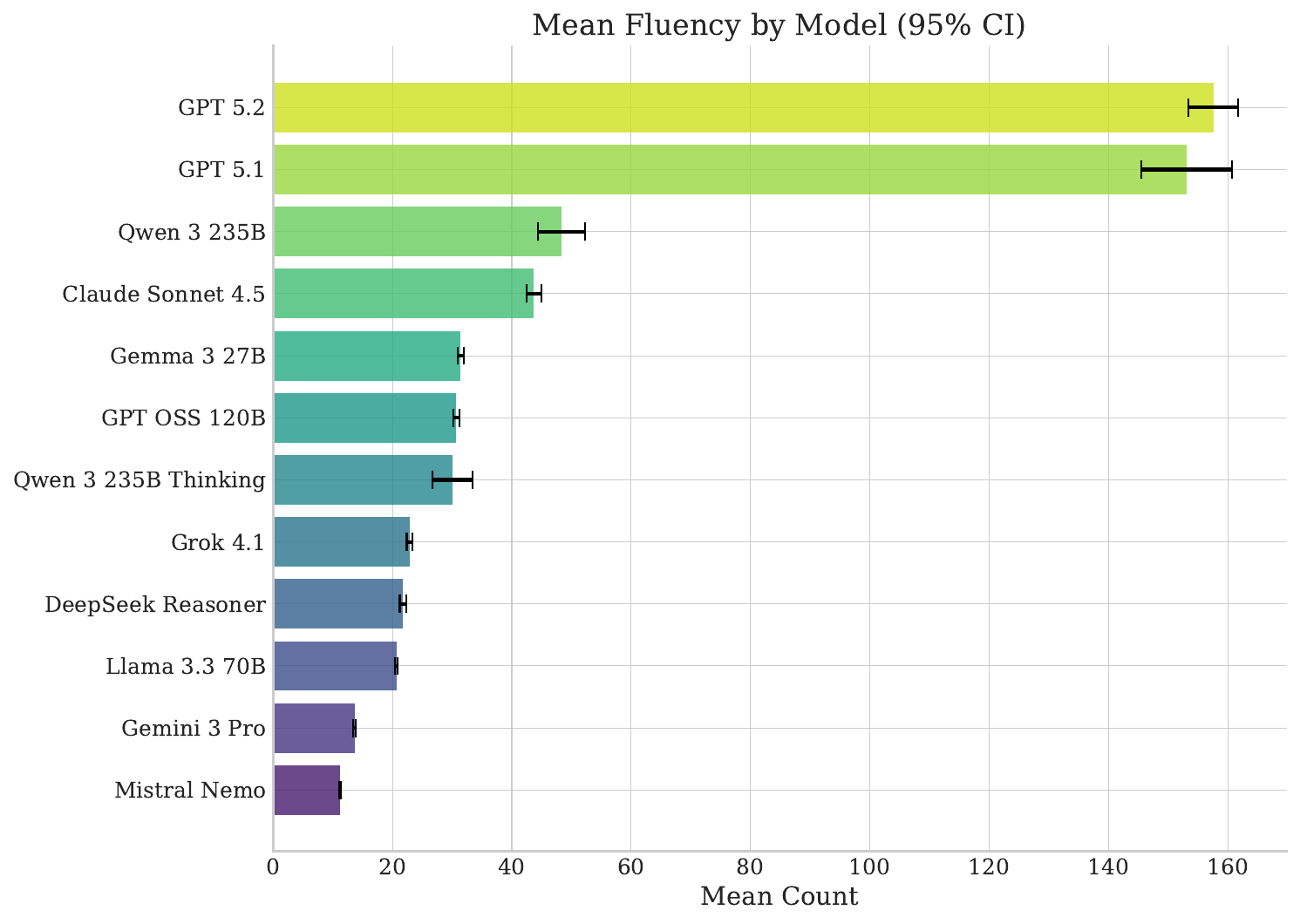}
        \caption{Mean Fluency by Model (95\% CI).}
        \label{fig:model_flu}
    \end{minipage}
\end{figure*}

\begin{figure*}[ht]
    \centering
    \includegraphics[width=0.6\textwidth]{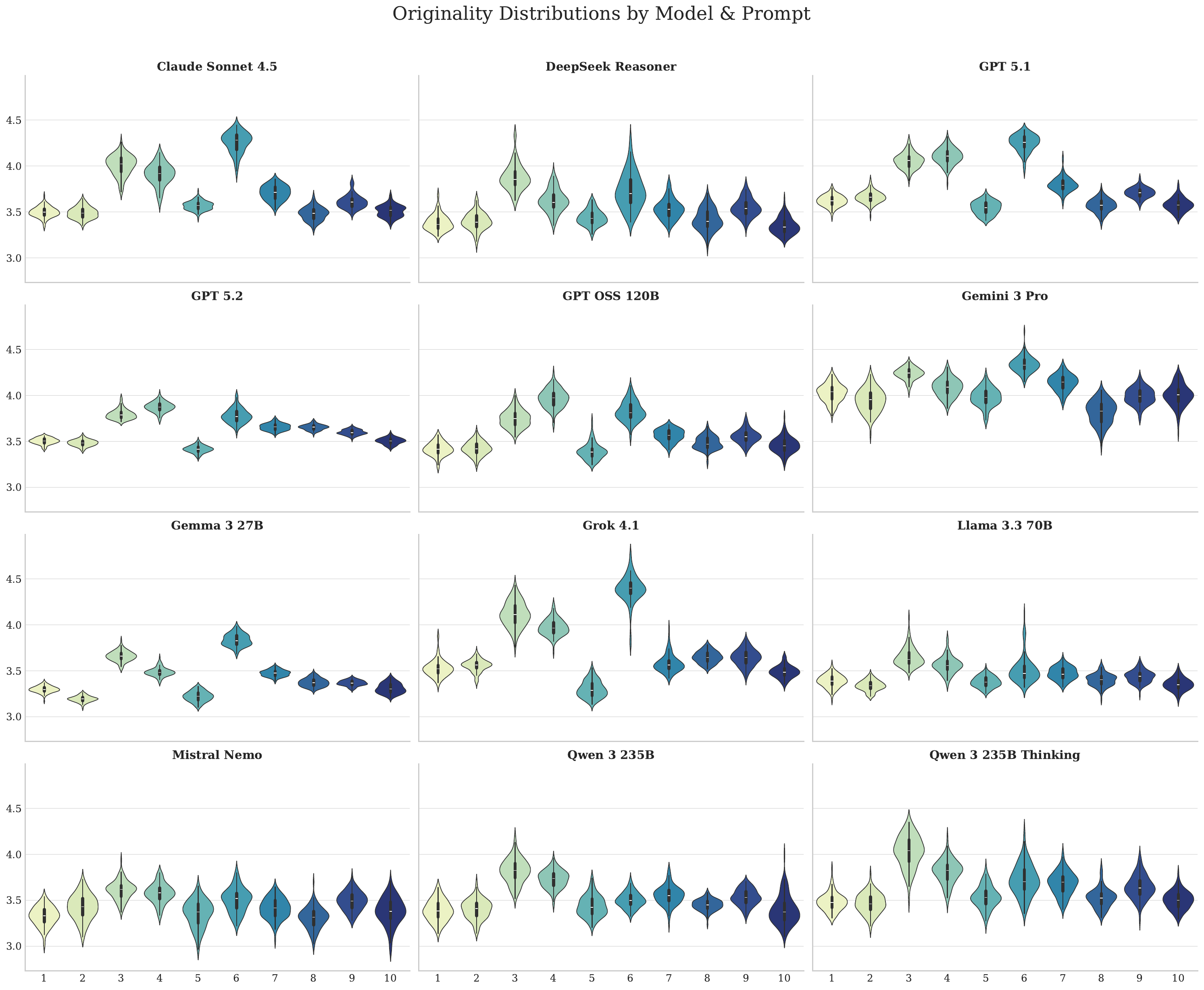}
    \caption{Violinplots displaying originality scores as a function of prompt separately for each model. \textbf{Prompt Key}: \textbf{P1}: Direct Baseline (Think of uses...). \textbf{P2}: One-Shot Example (Example: Use for yarn storage...). \textbf{P3}: Heuristic/Domain (Think across domains: art, survival...). \textbf{P4}: Anticipatory (Avoid generic ideas...). \textbf{P5}: Chain-of-Thought (Think step-by-step...). \textbf{P6}: Creative Persona (You are the most creative person...). \textbf{P7}: Phrasing Variation (Synonymous rewording). \textbf{P8}: Format Constraint (No titles or colons...). \textbf{P9}: Info Order (Emphasize inventive...). \textbf{P10}: Typo Robustness (Injected noise).}
    \label{fig:originality_faceted}
\end{figure*}

\begin{figure*}[ht]
    \centering
    \includegraphics[width=0.6\textwidth]{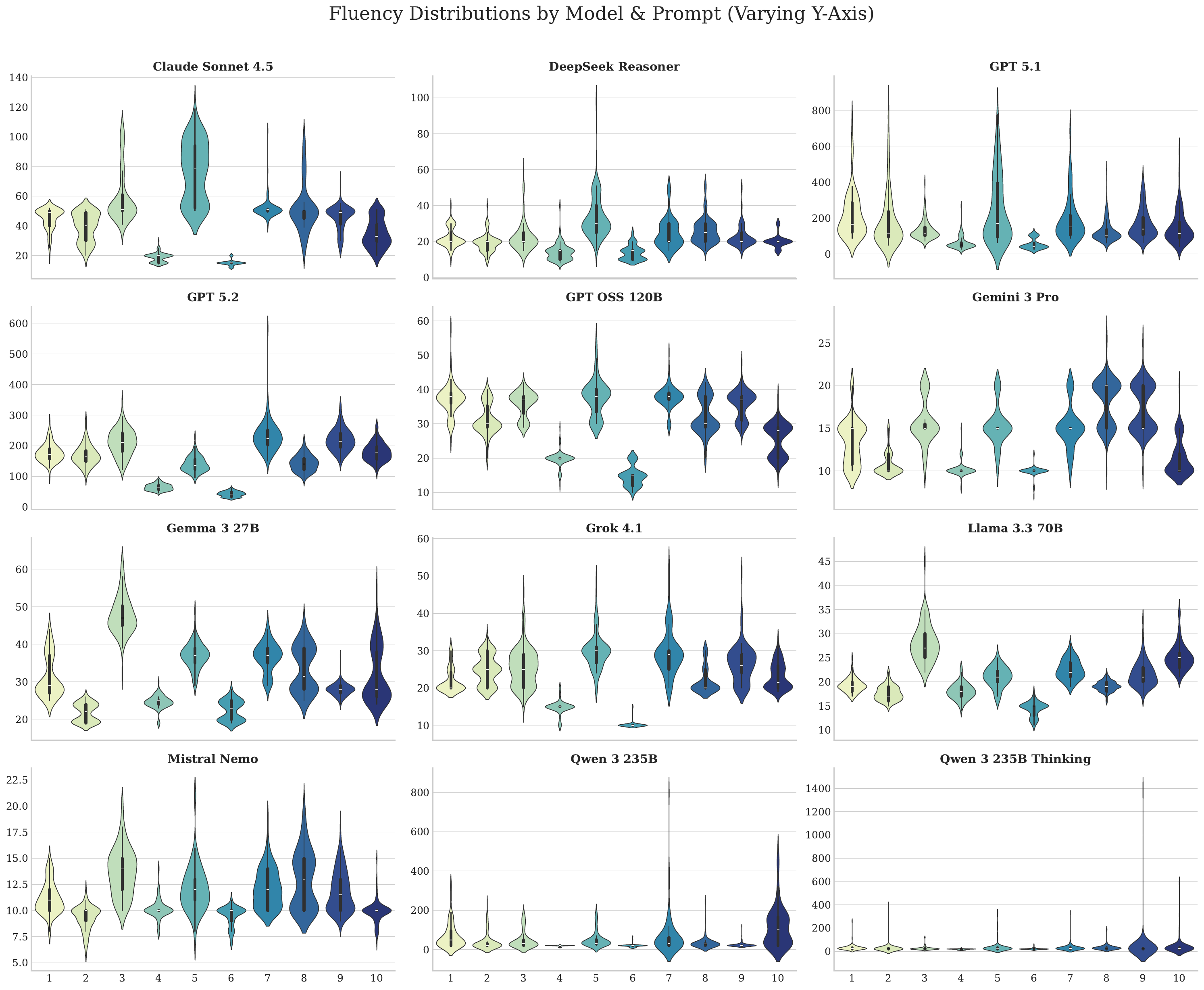}
    \caption{Violinplots displaying fluency (number of ideas) as a function of prompt separately for each model (varying y-axis range to show distribution detail). \textbf{Prompt Key}: \textbf{P1}: Direct Baseline (Think of uses...). \textbf{P2}: One-Shot Example (Example: Use for yarn storage...). \textbf{P3}: Heuristic/Domain (Think across domains: art, survival...). \textbf{P4}: Anticipatory (Avoid generic ideas...). \textbf{P5}: Chain-of-Thought (Think step-by-step...). \textbf{P6}: Creative Persona (You are the most creative person...). \textbf{P7}: Phrasing Variation (Synonymous rewording). \textbf{P8}: Format Constraint (No titles or colons...). \textbf{P9}: Info Order (Emphasize inventive...). \textbf{P10}: Typo Robustness (Injected noise).}
    \label{fig:fluency_varying}
\end{figure*}

\begin{figure*}[ht]
    \centering
    \includegraphics[width=0.6\textwidth]{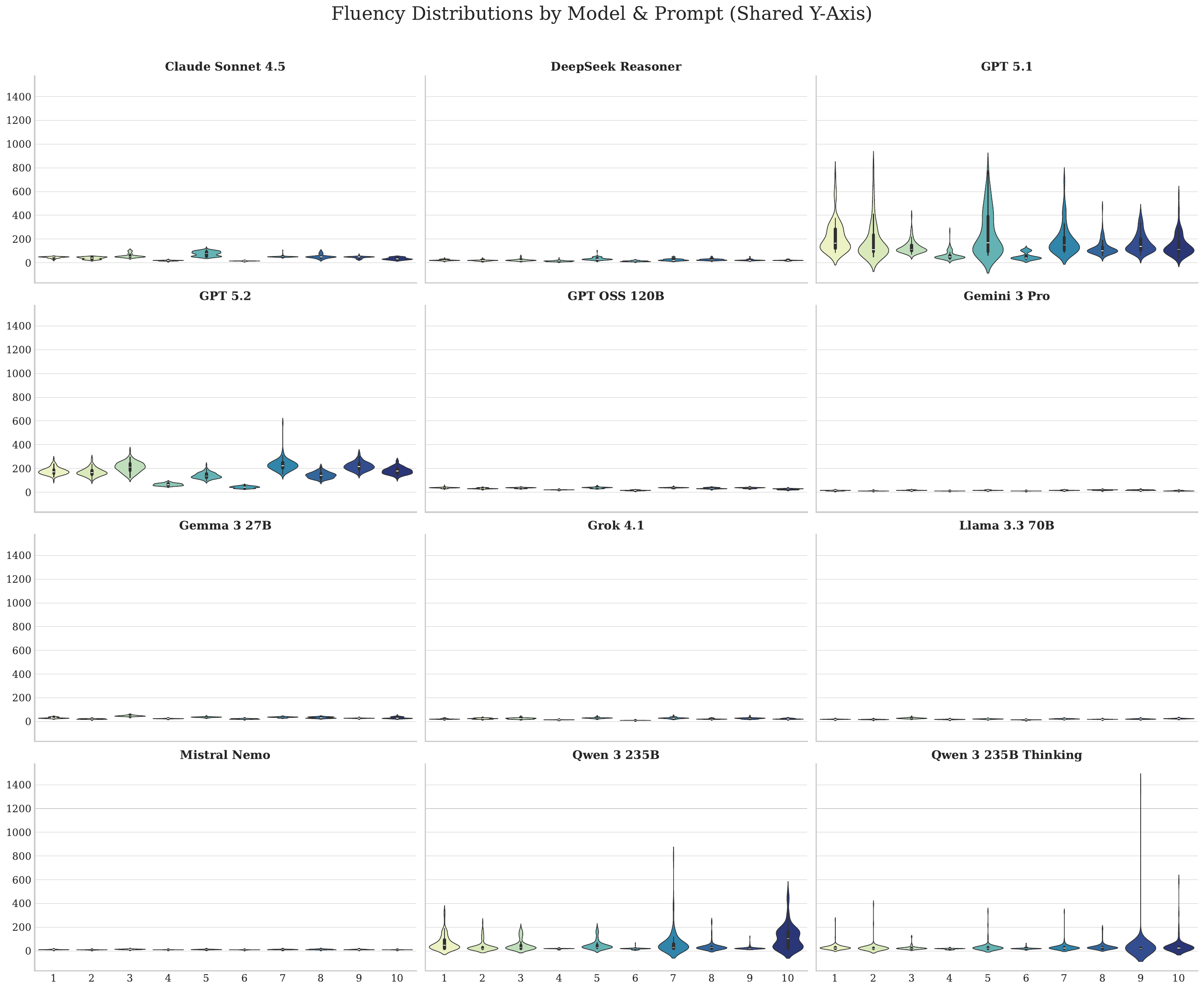}
    \caption{Violinplots displaying fluency (number of ideas) as a function of prompt separately for each model (shared y-axis to compare scales across models). \textbf{Prompt Key}: \textbf{P1}: Direct Baseline (Think of uses...). \textbf{P2}: One-Shot Example (Example: Use for yarn storage...). \textbf{P3}: Heuristic/Domain (Think across domains: art, survival...). \textbf{P4}: Anticipatory (Avoid generic ideas...). \textbf{P5}: Chain-of-Thought (Think step-by-step...). \textbf{P6}: Creative Persona (You are the most creative person...). \textbf{P7}: Phrasing Variation (Synonymous rewording). \textbf{P8}: Format Constraint (No titles or colons...). \textbf{P9}: Info Order (Emphasize inventive...). \textbf{P10}: Typo Robustness (Injected noise).}
    \label{fig:fluency_same}
\end{figure*}

\clearpage
\subsection{Specific Prompt Analyses}

\begin{figure*}[ht]
    \centering
    \includegraphics[width=0.9\textwidth]{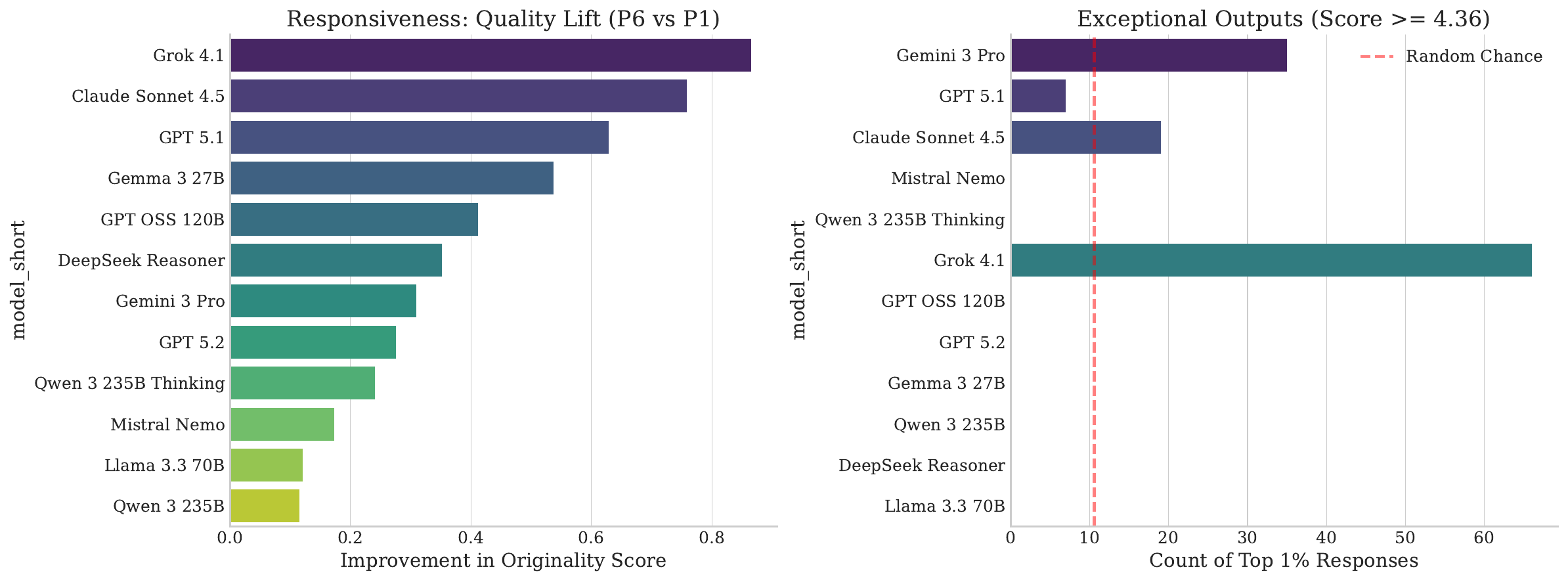}
    \caption{Distribution of Exceptional Responses (Top 1\% Originality) and Performance on Prompt 6. Grok 4.1 produces a disproportionate number of "outlier" high-quality ideas, particularly on the Persona prompt (P6).}
    \label{fig:outliers_prompt6}
\end{figure*}

\begin{figure*}[ht]
    \centering
    \includegraphics[width=0.9\textwidth]{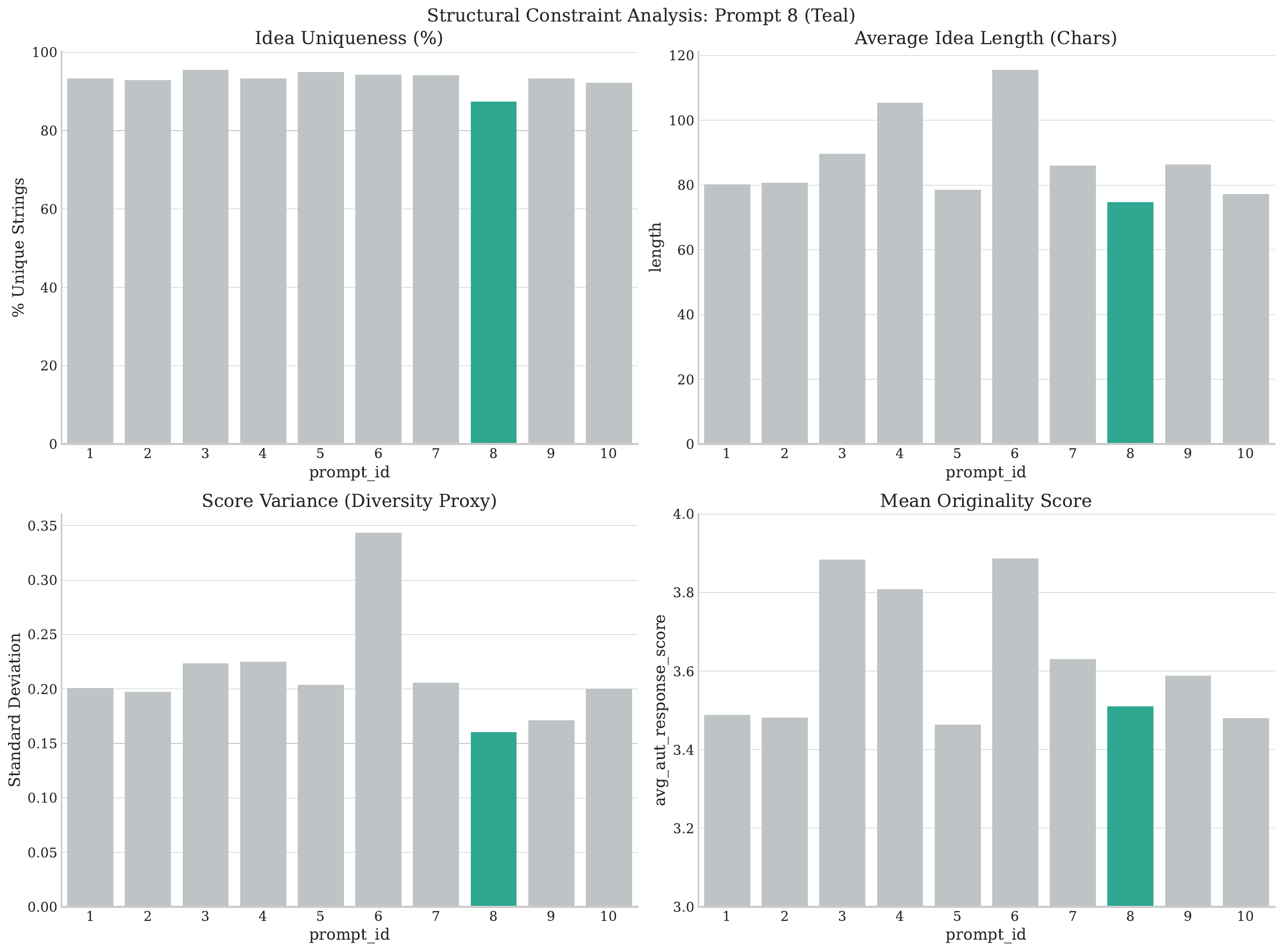}
    \caption{The "Structural Constraint" Effect. Prompt 8 (highlighted in teal) shows significantly lower semantic diversity and uniqueness compared to all other prompts due to its restrictive formatting instructions.
    \textbf{Prompt Key}: \textbf{P1}: Direct Baseline (Think of uses...). \textbf{P2}: One-Shot Example (Example: Use for yarn storage...). \textbf{P3}: Heuristic/Domain (Think across domains: art, survival...). \textbf{P4}: Anticipatory (Avoid generic ideas...). \textbf{P5}: Chain-of-Thought (Think step-by-step...). \textbf{P6}: Creative Persona (You are the most creative person...). \textbf{P7}: Phrasing Variation (Synonymous rewording). \textbf{P8}: Format Constraint (No titles or colons...). \textbf{P9}: Info Order (Emphasize inventive...). \textbf{P10}: Typo Robustness (Injected noise).}
    \label{fig:prompt8}
\end{figure*}

%%%%%%%%%%%%%%%%%%%%%%%%%%%%%%%%%%%%%%%%%%%%%%%%%%%%%%%%%%%%%%%%%%%%%%%%%%%%%%%
%%%%%%%%%%%%%%%%%%%%%%%%%%%%%%%%%%%%%%%%%%%%%%%%%%%%%%%%%%%%%%%%%%%%%%%%%%%%%%%

\end{document}